\documentclass{article}
\usepackage{spconf,amsmath}
\usepackage[pdftex]{graphicx}
\usepackage{color}
\usepackage{xspace}
\DeclareGraphicsExtensions{.pdf,.jpeg,.png,.jpg}
\usepackage[a4paper,pagebackref,hyperindex=true,hyperfootnotes=false]{hyperref}
\graphicspath{./figures/}
\usepackage{soul}
\usepackage{color, ifpdf}
\usepackage{comment}
\usepackage{authblk}
\newcommand\Mark[1]{\textsuperscript#1}

\usepackage[export]{adjustbox}
\usepackage{caption}
\usepackage{subcaption}
\usepackage{enumerate}
\usepackage{cite}
\usepackage{BOONDOX-cal}
\newcommand{\comments}[1]{}

\makeatletter
\DeclareRobustCommand\onedot{\futurelet\@let@token\@onedot}
\def\@onedot{\ifx\@let@token.\else.\null\fi\xspace}
\makeatother

\def\stateoftheart{state-of-the-art}
\def\stateofart{state-of-art}
\def\groundtruth{ground truth}
\def\eg{\emph{e.g}\onedot}

 \def\vs{\emph{vs}\onedot}
 
\def\etal{\emph{et al}\onedot}

\def\matlab{MATLAB}

\newcommand{\refsec}[1]{Section~\ref{#1}}
\newcommand{\reffig}[1]{Figure~\ref{#1}}

\newcommand{\reftab}[1]{Table~\ref{#1}}

\newcommand\blfootnote[1]{%
  \begingroup
  \renewcommand\thefootnote{}\footnote{#1}%
  \addtocounter{footnote}{-1}%
  \endgroup
}

\newcommand{\mycomment}[1]{}

\newcommand{\argmin}{\operatornamewithlimits{argmin}}
\title{Sub-cortical brain structure segmentation using F-CNN's}
\makeatletter
\def\@name{$^{*}$Mahsa Shakeri$^{2,4}$, $^{*}$Stavros Tsogkas\Mark{1}, Enzo Ferrante\Mark{1}, Sarah Lippe$^{3,4}$, \\Samuel Kadoury$^{2,4}$, Nikos Paragios\Mark{1}, Iasonas Kokkinos\Mark{1}}
\address{$^1$CVN, CentraleSupelec, Inria, Universite Paris-Saclay, $^2$ Polytechnique Montreal, \\ $^3$ University of Montreal, $^4$ Sainte-Justine Hospital Research Center}

\begin{document}
%\ninept
%
\maketitle

\begin{abstract}
In this paper we propose a deep learning approach for segmenting sub-cortical structures of the human brain in Magnetic Resonance (MR) image data. We draw inspiration from a \stateoftheart\ Fully-Convolutional Neural Network (F-CNN) architecture for semantic segmentation of objects in natural images, and adapt it to our  task. Unlike previous CNN-based methods that operate on image patches, our model is applied on a full blown 2D image, without any alignment or registration steps at testing time. We further improve segmentation results by interpreting the CNN output as potentials of a Markov Random Field (MRF), whose topology corresponds to a volumetric grid. Alpha-expansion is used to perform approximate inference imposing spatial volumetric homogeneity to the CNN priors. We compare the performance of the proposed pipeline with a similar system using Random Forest-based priors, as well as \stateofart\ segmentation algorithms, and show promising results on two different brain MRI datasets.\blfootnote{$^*$Authors contributed equally}
\end{abstract}
%
% ======================================================================
\section{Introduction}\label{sec:intro}
% ======================================================================
Image segmentation is a fundamental process in several medical applications. Diagnosis, treatment, planning and monitoring, as well as pathology characterization, benefit from accurate segmentation. In this paper we are interested in brain sub-cortical structures located at the frontostriatal system. Previous studies have shown the involvement of the frontostriatal structures in different neurodegenerative and neuropsychiatric disorders, including schizophrenia, Alzheimer’s disease, attention deficit, and subtypes of epilepsy~\cite{Chudasama2006frontostriatal}. Segmenting these parts of the brain enables a physician to extract various volumetric and morphological indicators, facilitating the quantitative analysis and characterization of several neurological diseases and their evolution. 

In the past few years, deep learning techniques, and particularly Convolutional Neural Networks (CNNs), have rapidly become the tool of choice for tackling challenging computer vision tasks. CNNs were popularized by Lecun, after delivering \stateofart\ results on hand-written digit recognition~\cite{lecun1998gradient}. However, they fell out of favor in the following years, mostly due to hardware and training data limitations. Nowadays, the availability of large-scale datasets (\eg ImageNet), powerful GPUs and appropriate software libraries, have rekindled the interest in deep learning and have made it possible to harness their power. Krizhevsky \etal \cite{krizhevsky2012imagenet} published results demonstrating clear superiority of deep architectures over hand-crafted features or shallow networks, for the task of image classification. Since then, CNNs have helped set new performance records for many other tasks; object detection, texture recognition and object semantic segmentation just to name a few.
% I removed this reference to gain space: long2014fully, object detection~\cite{girshick2014rich}, texture recognition~\cite{cimpoi2015deep}, and object semantic segmentation~\cite{chen2014semantic}, just to name a few.

Our work is similar in spirit to~\cite{prasoon2013deep}, but with some notable differences. In~\cite{prasoon2013deep} the authors train one CNN for each of the three orthogonal views of MRI scans, for knee cartilage segmentation, with the loss being computed on the concatenated outputs of the three networks. The inputs to each CNN are $28\times 28$ image patches and the output is a softmax probability of the central pixel belonging to the tibial articular cartilage. In contrast, our method operates on full 2D image slices, exploiting context information to accurately segment regions of interest in the brain. In addition, we use \emph{fully convolutional} CNNs~\cite{long2014fully} to construct dense segmentation maps for the whole image, instead of classifying individual patches. Furthermore, our method handles multiple class labels instead of delivering a foreground-background segmentation, and it does that efficiently, performing a single forward pass in$~5ms$.

CNNs are characterized by large receptive fields that allow us to exploit context information across the spatial plane. Processing 2D slices individually, however, means that we remain agnostic to \emph{3D context} which is important, since we are dealing with volumetric data. The obvious approach of operating directly on the 3D volume instead of 2D slices, would drastically reduce the amount of data available for training, making our system prone to overfitting, while increasing its computational requirements. Alternatively, we construct a Markov Random Field on top of the CNN output in order to impose volumetric homogeneity to the final results. The CNN scores are considered as unary potentials of a multi-label energy minimization problem, where spatial homogeneity is propagated through the pair-wise relations of a 6-neighborhood grid. For inference we choose the popular alpha-expansion technique that leads to guaranteed optimality bounds for the type of energies we define~\cite{boy01}.

% ======================================================================
\section{Using CNNs for Semantic Segmentation}\label{sec:cnns}
% ======================================================================
Our network is inspired by the Deeplab architecture that was recently proposed for semantic segmentation of objects~\cite{chen2014semantic}. Due to limited space, we refer the reader to~\cite{chen2014semantic} for details. One obvious and straightforward choice for adapting the Deeplab network to our task, would be to simply fine-tune the last three convolutional layers that replace their fully connected counterparts in the VGG-16 network, while initializing the rest of the weights to the VGG-16 values. This is a common approach when adapting an already existing architecture to a new task, but given the very different nature of natural RGB images and MR image data (RGB \vs grayscale, varying \vs black background), we decided to train a fully convolutional network from scratch.

Training a deep network from scratch presents us with some challenges. Medical image datasets tend to be smaller than natural image datasets, and segmentation annotations are generally hard to obtain. In our case, we only have a few 3D scans at our disposal, which increases the risk of overfitting. In addition, the repeated pooling and sub-sampling steps that are applied in the input images as it flows through a CNN network, decrease the output resolution, making it difficult to detect and segment finer structures in the human brain. To address these challenges, we make a series of design choices for our network: first, we opt for a shallower network, composed of five pairs of convolutional/max pooling layers. We sub-sample the input only for the first two max-pooling layers, and keep a stride of $1$ for the remaining layers, introducing holes, as in~\cite{chen2014semantic}. This allows us to keep increasing the effective receptive field of filters, without further reducing the resolution of the output response maps. For a $256\times256$ input image, the total sub-sampling factor of the network is $4$, resulting in a $64\times 64\times L$ array, where $L$ is the number of class labels. A $1-$pixel stride is used for all convolutional layers and $0.5$ activation probability for all dropout layers. The complete list of layers and important parameters is given in~\reftab{tab:architecture}.
At test time, a 2D image is fed to the network and the output is a three-dimensional array of probability maps (one for each class), obtained via a softmax operation. To obtain a brain segmentation at this stage, we simply resize the output to the input image dimensions using bilinear interpolation and assign at each pixel the label with the highest probability. However, we still need to impose volumetric homogeneity to the solution. We propose to do it using Markov Random Fields.

\begin{table}[t!]
\resizebox{\linewidth}{!}{
    \begin{tabular}{|c|c|c|c|c|c|c|}
    \hline
    Block    & conv kernel     & \# filters     & hole stride  & pool kernel   & pool stride   & dropout\\\hline
	1         & 7$\times$7      & 64           & 1				& 3$\times 3$   & 2      		& no     \\\hline
	2     	 & 5$\times$5     & 128          & 1  			& 3$\times 3$   & 2      		& no     \\\hline
	3         & 3$\times$3      & 256          & 2				& 3$\times 3$   & 1       		& yes    \\\hline
	4         & 3$\times$3      & 512          & 2  			& 3$\times 3$   & 1       		& yes    \\\hline
	5         & 3$\times$3      & 512          & 2				& 3$\times 3$   & 1       		& yes    \\\hline
	6         & 4$\times$4      & 1024         & 4				& no pooling    &        		& yes    \\\hline
	7         & 1$\times$1      & 39           & 1				& no pooling    &        		& no     \\\hline
    \end{tabular}
    }
    \caption{Layers used in our architecture. All convolutional layers have a stride of one pixel; a hole stride of "1" means that we introduce no holes.}
    \label{tab:architecture}
\end{table}

\subsection{Multi-label segmentation using CNN-based priors}\label{sec:segmentation}
For every slice of a 3D image, the output of the proposed CNN is a softmax map that indicates the probability of every pixel to be part of a given brain structure $l \in \mathcal{L}$ (label). We consider the volume $P^{\mathrm{CNN}}_{i}(l):\mathcal{L} \rightarrow [0,1] $ formed by the stacked CNN output slices, as a prior of the brain 3D structures, where $i$ indicated a voxel from the original image.

Let $\mathcal{G}=\langle \mathcal{V},\mathcal{E} \rangle$ be a graph representing a Markov Random Field, where nodes in $\mathcal{V}$ are variables (voxels) and $\mathcal{E}$ is a standard 6-neighborhood system defining a 3D grid. Variables $i \in \mathcal{V}$ can take labels $l_i$ from a labelspace $\mathcal{L}$. A labeling $\mathcal{S}= \{l_i \mid i \in \mathcal{V} \}$  assigns one label to every variable. We define the energy $E(\mathcal{S})$ which consists of unary potentials $V_i$ and pair-wise potentials $V_{ij}$ such that it is minimum when $\mathcal{S}$ corresponds to the best possible labeling.

Unary terms are defined as $V_i(l_i) = -\log(P^{\mathrm{CNN}}_{i}(l_i))$, and they assign low energy to high probability values. Pair-wise terms encode the spatial homogeneity constraint by simply encouraging neighbor variables to take the same semantic label. In order to align the segmentation boundaries with intensity edges, we made this term inversely proportional to the difference of the intensity $I_i$ and $I_j$ associated to the given voxels. The pair-wise formulation is $V_{i,j}(l_i, l_j) = w_{ij}.[l_i\neq~l_j]$ where $w_{ij} = \exp\left(-\frac{\mid (I_i - I_j) \mid^2}{2 \sigma^2}\right)$. Finally, the energy minimization problem is defined as:
\begin{equation}
\mathcal{S}^* = \argmin E(\mathcal{S}) =  \argmin \sum\limits_{i \in \mathcal{V}} V_i(l_i) + \lambda \hspace{-2mm} \sum\limits_{(i,j) \in \mathcal{E}} \hspace{-2mm} V_{i,j}(l_i, l_j).
\end{equation}
$\mathcal{S}^*$ represents the optimal label assignment. Note that this energy is a metric in the space of labels $\mathcal{L}$; thus, it is guaranteed that using alpha-expansion technique we can find a solution $\hat{\mathcal{S}}$  whose energy lies within a factor of 2 with respect to the optimal energy (i.e. $E(\hat{\mathcal{S}}) \leq 2.E({\mathcal{S}^*})$). Alpha-expansion is a well known move-making technique to perform approximate inference using graph cuts, that has shown to be accurate in a broad range of vision problems. We refer the reader to \cite{boy01} for a complete discussion on energy minimization using alpha-expansion.

% ======================================================================
\section{Experiments and Discussion}\label{sec:results}
% ======================================================================
\begin{figure*}[t!]
\centering
\includegraphics[width=\textwidth]{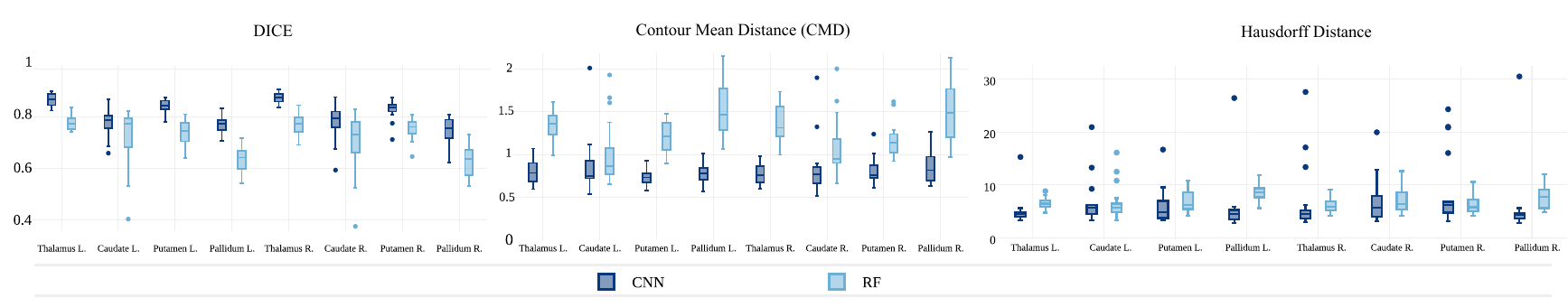}
\caption{Average Dice coefficient, Hausdorff distance, and contour mean distance on eight subcortical structures of IBSR dataset. The proposed CNN-based method outperforms the RF-based approach (better viewed in color and magnified).}
\label{fig:IBSR}
\end{figure*}

\begin{figure}[t!]
\includegraphics[scale=1.05]{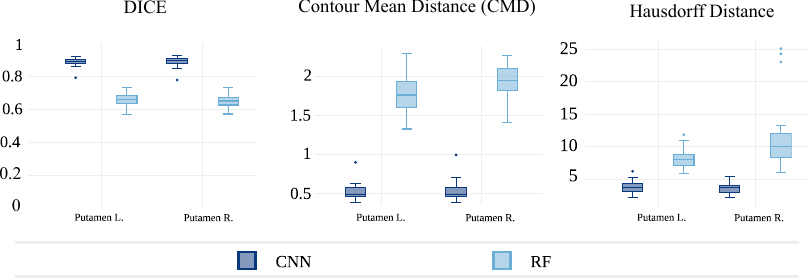}
\caption{The average Dice coefficient, Hausdorff distance, and contour mean distance on left and right putamen structure of RE dataset. The proposed CNN-based method generates more accurate segmentation results compared to the RF-based approach (better viewed in color and magnified).}
\label{fig:MTL}
\end{figure}

We used the proposed method to segment a group of sub-cortical structures located at the frontostriatal network, including thalamus, caudate, putamen and pallidum. We evaluated our approach on two brain MRI datasets.

The first one is a publicly available dataset provided by the Internet Brain Segmentation Repository (IBSR)~\cite{Rohlfing2012Image}. It contains $18$ labeled 3D T1-weighted MR scans with slice thickness of around $1.3~mm$. In this work we use the subset of $8$ primarily subcortical labels, including left and right thalamus, caudate, putamen, and pallidum. The second dataset is obtained from a Rolandic Epilepsy (RE) study, including $17$ children with epilepsy and $18$ matched healthy individuals. For each participant, T1-weighted magnetic resonance images (MRI) were acquired with a $3$ T scanner (Philips Acheiva) with an in-plane resolution of $256 \times 256$ and slice thickness of $1~mm$. The left and right putamen structures were manually annotated by an experienced user. For both datasets, we process volumes slice by slice, after resizing them to $256 \times 256$ pixels. We treat these 2D slices as individual grayscale images to train our CNN.

In the first experiment, we compare the performance of our segmentation method using CNN priors, with an approach based on Random Forest priors, where the same MRF refinement is applied. The RF-based per-voxel likelihoods are computed in the same way as~\cite{Alchatzidis2014Discrete}. Then, the RF probability maps are considered as the unary potentials of a Markov Random Field and alpha-expansion is used to compute the most likely label for each voxel, as explained in \refsec{sec:segmentation}.~\reffig{fig:IBSR} and~\reffig{fig:MTL} show the average Dice coefficient, Hausdorff distance, and contour mean distance between output segmentations and the \groundtruth\ for different structures. These results show that the CNN-based approach achieves higher Dice compared to RF-based method, while producing lower Hausdorff and contour mean distance. 

\begin{figure}[t!]
\includegraphics[scale=0.4]{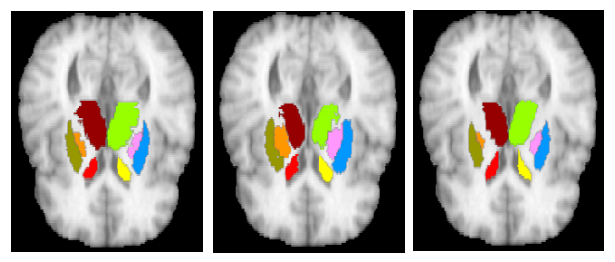}
\caption{2D slice segmentation (IBSR). \textbf{Left:} Groundtruth. \textbf{Middle:} RF-based results. \textbf{Right:} CNN-based results.}
\label{fig:visualRes}
\end{figure}

In the second experiment, we compare the accuracy of our proposed method with two publicly available \stateoftheart\ automatic segmentation toolboxes, Freesurfer~\cite{Fischl2002Whole}, and FSL-FIRST~\cite{Patenaude2011Bayesian}. In \reftab{tab:IBSRMTL} we report the average Dice coefficient for the left and right structures; these results show that our method provides better segmentations compared to the \stateoftheart\ for three sub-cortical structures in both IBSR and RE dataset. However, Freesurfer results in better segmentation for caudate in the IBSR dataset which could be attributed to the limitation of CNN in capturing thin tail areas of the caudate structures. In~\reffig{fig:visualRes} we show qualitative results.

\subsection{CNN Training and Evaluation Details}\label{sec:training}
The input to our network is a single 2D slice from a 3D MRI scan, along with the corresponding label map. We apply data augmentation to avoid overfitting: we use horizontally flipped and translated versions of the input images by 5, 10, 15, 20 pixels, across the $x/y$ axes. Other transformations, such as rotation, could be considered as well. The MR image data are centered and the background always takes zero values, so we do not perform mean image subtraction as is usually the case. 

In the case of IBSR, we split the available data into three sets. Each time, we use two of the sets as training data (approximately $100K$ training samples) and the third set as test data. One of the training data volumes is left out and used as validation data. Similarly, we split RE into two subsets of equal size, using one for training and one for testing, each time. We train on both datasets for $35$ epochs starting with a learning rate of $0.01$ and dropping it at a logarithmic rate until $0.0001$. For training, we use standard SGD with a momentum of $0.9$ and a softmax loss. For all our experiments we used \matlab\  and the deep learning library MatConvNet~\cite{vedaldi2014matconvnet}. Code, computed probability maps, and more results can be found at~\url{https://github.com/tsogkas/brainseg}.

We also experimented with CNNs trained on 2D slices from the other two views (sagittal and coronal) but the resulting models performed poorly. The problem is rooted in the inherent symmetry of some brain structures and the fact that the CNN is evaluated on individual slices, ignoring 3D structure. For instance, when processing slices across sagittal view, the right and left putamen appear at roughly the same positions in the image. They are also very similar in terms of shape and appearance, which fools the system into assigning the same label to both regions. This simple example demonstrates the need for richer priors that take into account the full volume structure to assign class labels.

% ======================================================================
\section{Conclusion}\label{sec:conclusion}
% ======================================================================
%\vspace{-1em}
In this paper, we proposed a deep learning framework for segmenting frontostriatal sub-cortical structures in MR images of the human brain. We trained a fully convolutional neural network for segmentation of 2D slices and treated the output probability maps as a proxy for the respective voxel likelihoods. We further improved segmentation results by using the CNN outputs as potentials of a Markov Random Field (MRF) to impose spatial volumetric homogeneity.
Our experiments show that the proposed method outperforms approaches based on other learned priors, as well as \stateoftheart\ segmentation methods.
However, we also note some limitations: the current model is not able to accurately capture thin tail areas of the caudate structures. Second, symmetric structures confound the CNN training process when considering views which are parallel to the plane of symmetry. Third, graph-based methods have to be used to impose volumetric consistency since training is done on 2D slices. Different network layouts, taking account of volumetric structure can possibly help overcome these limitations.

\begin{table}[t!]
\caption{The average Dice coefficient of the three methods on different brain structures. Values are reported as the average of the left and right structures.}
\tiny
\resizebox{\linewidth}{!}{
    \begin{tabular}{|c|c|c|c|c}
    \hline
	   & Proposed     & Freesurfer    & FSL \\ \hline
	IBSR-Thalamus      & \bf0.87     & 0.86  & 0.85 \\ \hline
	IBSR-Caudate     & 0.78     & \bf0.82   & 0.68 \\ \hline
	IBSR-Putamen     & \bf0.83   & 0.81  & 0.81 \\ \hline
	IBSR-Pallidum    & \bf0.75   & 0.71  & 0.73 \\ \hline
	RE-Putamen & \bf0.89 & 0.74 & 0.88 \\ \hline
	
\end{tabular}
    \label{tab:IBSRMTL}}
\end{table}

% ======================================================================
\bibliographystyle{IEEEbib}
\bibliography{isbi2016}
\end{document}